\documentclass[fleqn,10pt,twocolumn]{SWARM24}
\usepackage{amsmath}

\DeclareMathOperator{\sgn}{sgn}

\title{The Impact of Network Structure on Ant Colony Optimization}

\author{Taiyo  Shimizu${}^{1\dagger}$ and Shintaro Mori${}^{2}$}
\speaker{Taiyo Shimizu}

\affils{${}^{1}$Graduate School of Science and Technolony, Hirosaki University\\
(E-mail: h24ms110@hirosaki-u.ac.jp)\\
${}^{2}$Graduate School of Science and Technolony, Hirosaki University\\
(Tel: +81-172-39-3628; E-mail: shintaro.mori@hirosaki-u.ac.jp)\\
}
\abstract{%
Ant Colony Optimization (ACO) is a swarm intelligence methodology utilized for solving optimization problems through information 
transmission mediated by pheromones. As ants sequentially secrete pheromones that subsequently evaporate, the information conveyed 
predominantly comprises pheromones secreted by recent ants. This paper introduces a network structure into the information 
transmission process and examines its impact on optimization performance. The network structure is characterized by an 
asymmetric BA model with parameters for in-degree $r$ and asymmetry $\omega$. At $\omega=1$, the model describes a scale-free 
network; at $\omega=0$, a random network; and at $\omega=-1$, an extended lattice. We aim to solve the ground state search 
of the mean-field Ising model, employing a linear decision function for the ants with their response to pheromones quantified 
by the parameter $\alpha$. For $\omega>-1$, the pheromone rates for options converge to stable fixed points of the stochastic system. 
Below the critical threshold $\alpha_c$, there is one stable fixed point, while above $\alpha_c$, there are two. Notably, as $\omega \to -1$, 
both the driving force toward stable fixed points and the strength of the noise reach their maximum, significantly enhancing the 
probability of finding the ground state of the Ising model.
}

\begin{document}

\pagestyle{headings}

\maketitle


\section{Introduction}
Network science emerged as a research field in the 20th century and has been actively studied in various academic disciplines\cite{Barabashi:2016,Newman:2018}. 
Particularly, the discoveries of "small-world" properties by Watts and Strogatz in 1998\cite{Watts:1998} and "scale-free" properties by Barabási and Albert 
in 1999\cite{Barabashi:1999} demonstrated the essential role of network structures in understanding complex systems. For instance, small-world properties are 
associated with rapid information transmission within networks, while scale-free properties highlight the presence of highly connected 
hubs in networks, which is fundamentally related to the spread of infectious diseases and systemic risks in the Internet\cite{Barabashi:2016}.

On the other hand, Ant Colony Optimization (ACO) is a heuristic approach developed based on the collective foraging behavior of 
ants to solve optimization problems\cite{Dorigo:1992,Dorigo:1997}. Since its inception, ACO has been 
successfully applied to various optimization problems\cite{Dorigo:2010}. 
ACO belongs to the category of swarm intelligence\cite{Gad:2022}, and recently, it has been applied to the distributed control of robot swarms\cite{Dorigo:2013}. 
Here, the network structure of robot swarms is known to play a crucial role in the robustness of swarm functionality. 
Particularly, the existence of hubs is undesirable due to the heavy burden on hub robots and their contribution to the vulnerability of the entire swarm.

What kind of network structure for ants is preferable in Ant Colony Optimization? In ACO, ants share information with each other mediated by pheromones to construct solutions. As pheromones evaporate, ants tend to focus on the information from ants that recently secreted pheromones. 
The network structure of ant pheromone referencing can be described as an extended lattice structure with links between recent ants. 
However, it remains unclear whether the extended lattice structure is the most optimal network structure.

This study aims to elucidate the optimal pheromone referencing network structure in ACO. Specifically, we consider a network where pheromones are 
referenced from a random selection of $r$ past ants. When selecting randomly, we incorporate the mechanism of 
the asymmetric Barabási-Albert (BA) model for growing networks\cite{Hisakado:2021}. The asymmetric BA model has two parameters, the node's in-degree $r$ and the parameter of asymmetry $\omega$. Particularly, when $\omega=-1$, it corresponds to an extended lattice; when $\omega=0$, a random graph; and when $\omega=1$, a BA model, resulting in a scale-free degree distribution with the existence of hubs\cite{Barabashi:1999}. Increasing $\omega$ further highlights the hub-and-spoke structure while maintaining scale-free properties. Conversely, when $\omega$ approaches $-1$, the degree distribution is similar to the extended lattice, but the node-to-node distance exhibits a small-world property\cite{Hisakado:2021}. However, the clustering coefficient is very small, indicating that it does not strictly possess small-world characteristics. We investigate the influence of the network structure on the ground state search of the Ising model\cite{Mori:2024-2} as affected by $\omega$ in the pheromone referencing network.

The structure of the paper is as follows: In Section 2, we introduce an ACO model with a pheromone-referencing network incorporating an asymmetric BA model for the ground state search of the Ising model. We define the ant's decision function as a linear function of pheromone ratio and describe the response to pheromone concentration with parameter $\alpha$. In Section 3, we describe the model using mean-field approximation with multi-dimensional stochastic differential equations (SDEs). Particularly, for $\omega>-1$, the drift and diffusion terms decrease with time, and in the limit of $t\to\infty$, the pheromone ratio converges to stable fixed points of the stochastic system. The number of stable fixed points changes with $\alpha$ and exceeds the threshold $\alpha_c$, indicating a discontinuous transition. For $\omega=-1$, the stochastic differential equations exhibit a structure similar to when pheromones evaporate, and in the limit of $t\to\infty$, the pheromone ratio follows a stationary distribution of the Fokker-Planck equation. 
We also  validate the theory with numerical simulations and further investigate the relationship between the ground state search probability and the parameter $\omega$ determining the network structure in section 4. 
Section 5 summarizes our findings.

\section{Asymmetric BA model for Pheromone reference network and ACO}

We tackle the problem of identifying the ground state of the infinite-range Ising model or the complete graph Ising model, characterized by $N$ binary variables $X(i) \in {0,1}$\cite{Stanley:1971,Nishmori:2001}. The system's energy is defined as:
\begin{eqnarray}
&&E[{X(i)}] = -\sum_{i}h(2X(i)-1) \nonumber \\
&-& \frac{1}{N-1}\sum_{i,j,i\neq j} J(2X(i)-1)(2X(j)-1)\nonumber .
\end{eqnarray}
Here, $J$ represents the exchange interaction strength, and $h$ denotes the external field. The transformation $\sigma(i)=2X(i)-1$ maps the binary variables $X(i)\in {0,1}$ to Ising spin variables $\sigma(i) \in {\pm 1}$. The system's energy is written with $\{\sigma(i)\}$ as,
\[
E[{\sigma(i)}] = -\sum_{i}h\sigma(i)-\frac{1}{N-1}\sum_{i,j,i\neq j} J\sigma(i)\sigma(j).
\]
The ground state for $h>0$ is uniformly $X(i)=1(\sigma(i)=1)$, 
with the energy being $-N(J+h)$. At $h=0$, two ground states exist with the ground state energy $-NJ$: $X(i)=1(\sigma(i)=1)$ for all $i$ and $X(i)=0(\sigma(i)=-1)$ for all $i$. The external field breaks the degeneracy, and the energy difference between these states for $h \neq 0$ is $2Nh$. The energy, given the magnetization $m = \sum_{i}(2X(i)-1)/N=\sum_{i}\sigma(i)/N$, is $E(m)=-N(hm+Jm^2)$. The energy barrier from $m=-1$ to $m=1$ is $E(0)-E(-1)=N(J-h)$, making the discovery of the ground state ($m=1$) challenging if $m=-1$ is initially found, especially when $J \gg 1$ and $h \ll 1, h > 0$.

Ants sequentially search for the ground state of the Ising model\cite{Mori:2024}. The choice made by the $t$th ant for $X(i)$ is denoted as $X(i,t) \in {0,1}$. In ACO, multiple ants typically search for the optimal solution simultaneously in each iteration. However, in this model, only one ant conducts the search in each iteration. The evaluation of the choice ${X(i,t)}, i=1,\cdots,N$ is based on the energy value, denoted as $E(t) = E[{X(i,t)}]$. Ant $t$ deposits pheromones on their choices $X(n,t) \in {0,1}$, with the amount of pheromone given by the Boltzmann weight $e^{-E(t)}$.

\begin{figure}[htbp]
\begin{center}
\includegraphics[width=7cm]{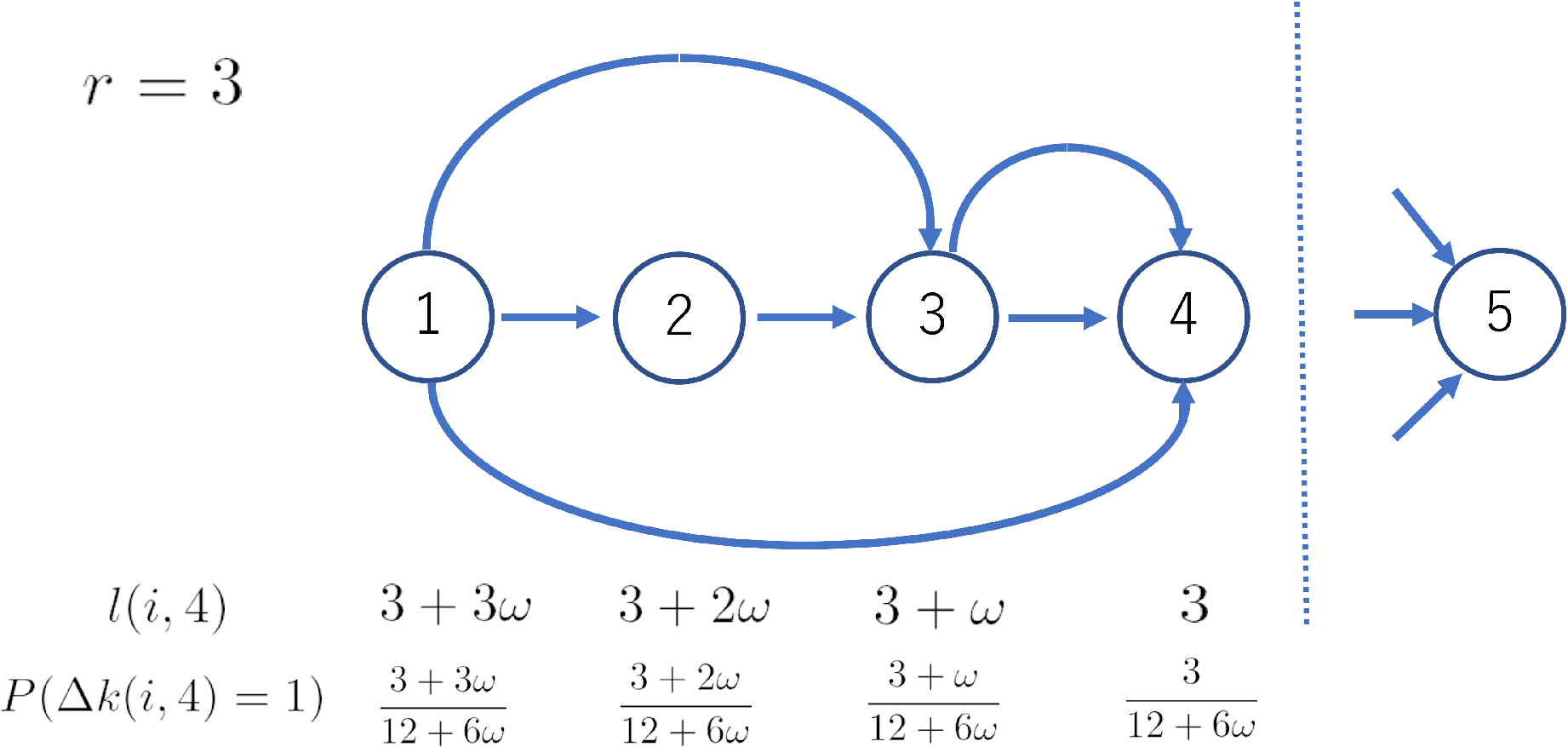}
\caption{\label{fig1}Initial configuration of the network for $r=3$ and $t=4$.}
\end{center}
\end{figure}

The first $r+1$ ants refer to all previous ants and they construct the complete graph. 
Figure \ref{fig1} shows the initial configuration of the network for $r=3$ and $t=4$.
Ant $t+1 \geq r+2$ randomly selects different $r$ ants from the $t\ge r+1$ ants that have already 
provided their solution. The probability that ant $i$ is chosen by ant $t+1$ is 
proportional to the popularity of ant $i$, denoted as $l(i,t)$\cite{Hisakado:2021}. $l(i,t)$ is defined as: 
\begin{equation}
l(i,t)=r+\omega k^{OUT}(i,t).
\end{equation}
Here, $k^{OUT}(i,t)$ represents the out-degree of the pheromone reference network after 
ant $t$ has chosen $r$ ants. At $t=r+1$, $k^{OUT}(i,r+1)=r+1-i$ for $i\leq r+1$, as the 
first $r+1$ ants form the complete graph. We choose $\omega \geq -1$.
Even for $\omega=-1$, ants can choose different $r$ ants.
In general, for $\omega\ge -1$, ants can choose different $r$.
If ant $i$ is chosen by ant $t+1$, the out-degree of ant $i$ increases by 1:
\[
\Delta k^{OUT}(i,t)=k^{OUT}(i,t+1)-k^{OUT}(i,t)=1.
\]
We assume that ant $i$ is never chosen by ant $t+1$ if $l(i,t)<0$. 
The maximal number of the out-degree of ant $i$ for $\omega<0$ is:
\[
k^{OUT}(i,t)\le \left\lceil \frac{r}{\omega} \right\rceil \,\, \mbox{for} \omega <0.
\]
Here, $\lceil x \rceil$ denotes the ceiling function. 
For $\omega \geq 0$, there is no limit on the maximal number of out-degrees. 
The probability that ant $i$ is chosen by ant $t+1$ is:
\[
P(\Delta k^{OUT}(i,t)=1)=r\frac{\mbox{Max}(l(i,t),0)}{\sum_{i=1}^{t}\mbox{Max}(l(i,t),0)}.
\]

\begin{figure}[htbp]
\begin{center}
\includegraphics[width=5.5cm]{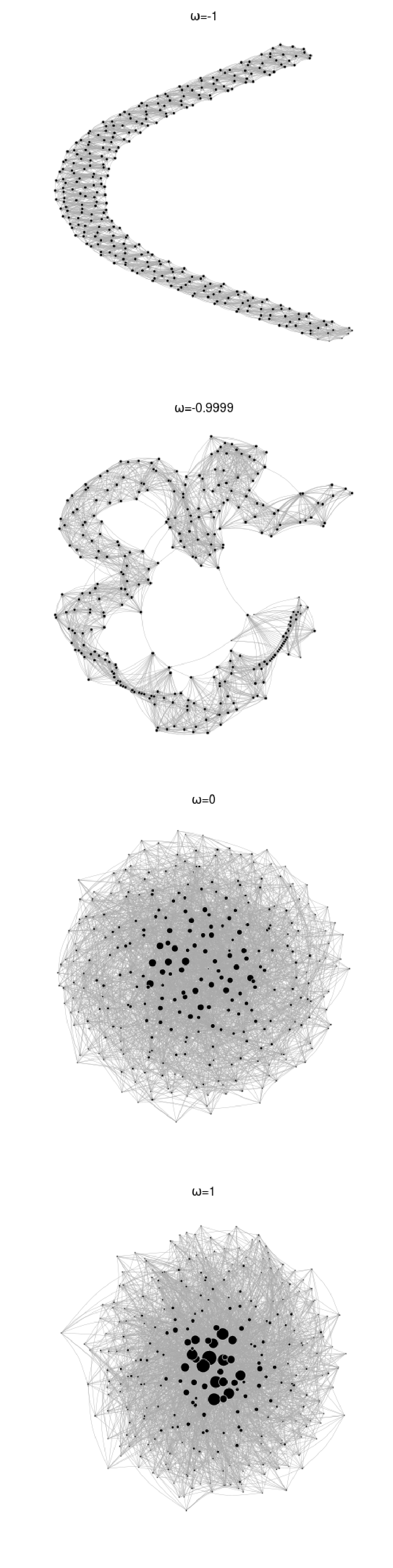}
\caption{\label{fig2}Typical network configuration for $r=10$ and $t=300$.
$\omega\in \{-1,-0.9999,0,1\}$.}
\end{center}
\end{figure}

In the case $\omega=-1$, ant $t+1$ chooses the previous $r$ ants, and 
$\Delta k^{OUT}(i,t)=1$ for $i=t\geq r+1$. 
The pheromone reference network becomes the extended 1-dimensional lattice. 
The pheromone does not evaporate, and ant $t+1$ references the previous $r$ ants; 
$r$ plays the role of pheromone evaporation. 
When $\omega>-1$, the extended lattice structure is broken and the mean distance 
between the ants becomes extremely short. At $\omega=0$, the ants 
choose the previous $r$ ants completely random and the network becomes random graph.
For $\omega>0$, the distribution of the out-degrees obeys power-law behavior.
Figure \ref{fig2} shows the typical configuration of the network for $r=10$ and $t=300$.
The sizes of the nodes is proportional to the out-degree. At $\omega=-1$, one can 
see the extended lattice structure. At $\omega=-0.9999$, the connection between 
distant nodes appears and the network becomes small world in the sense of 
average node-to-node distance. For $\omega=1$, the network has hubs with large 
out-degrees, which are represented by big black circles.

The first $r+1$ ants does not observe the vales of the pheromones and 
they decide by themselves.
Ant $t+1\ge r+2$ observes the values of the pheromones of $r$ ants linked 
in the pheromone reference network. The total value of pheromones 
of the $r$ ants is:
\begin{equation}
S(t)=\sum_{s=1}^t e^{-E(s)}\Delta k^{OUT}(s,t).
\label{eq:S}
\end{equation}
The pheromone on the choice $X(k,s)=x,1\leq s \leq t$ is:
\begin{equation}
S_x(k,t)=\sum_{s=1}^{t}e^{-E(s)}\Delta k^{OUT}(s,t)\delta_{X(k,s),x}.
\label{eq:Sx}
\end{equation}
Here, $\delta_{x,y}$ represents the Kronecker delta function, 
which is defined to be 1 if $x=y$ and 0 otherwise.

Ant $t+1\ge r+2$ makes decisions based on simple probabilistic rules. 
The information provided by $S_x(k,t)$ gives ant $t+1$ an indirect 
clue about the choice. If $S_1(k,t)>S_0(k,t)$, the posterior probability 
for $X(k,t+1)=1$ is larger than $1/2$ in Bayesian statistics. 
We denote the ratio of the pheromones on the choice $X(k,s)=1,s=1,\cdots,t$ 
in the pheromone reference network as $Z(k,t)$:
\begin{equation}
Z(k,t)\equiv \frac{S_1(k,t)}{S(t)}.
\label{eq:Z}
\end{equation}
The probability of the choice $X(k,t+1)=1$ is expressed as:
\begin{equation}
P(X(k,t+1)=1)=(1-\alpha)\frac{1}{2}+\alpha Z(k,t)\equiv f(Z(k,t)).
\label{eq:Px}
\end{equation}
Here, we introduce a decision function $f(z)$:
\[
f(z)\equiv (1-\alpha)\frac{1}{2}+\alpha z  \label{eq:f}.
\]

The parameter $0\leq \alpha<1$ determines the response of the choice to the 
values of the pheromones. $S_1(k,t)=0$ and $S_1(k,t)=S(t)$ are the absorbing 
states for $\alpha=1$; we restrict $\alpha<1$. When $\alpha=0$, $P(X(k,t+1)=1)=1/2$, 
and the ants choose at random. As $\alpha$ increases, the ants take into account 
the pheromone in their decisions. In ACO, the decision function adopts a nonlinear 
form $\propto S_1(k,t)^{\alpha}$. In the binary choice case, the decision under 
the case $Z(k,t)=S_1(k,t)\simeq S(t)/2$ is crucial. The above linear form approximates 
the usual decision function in the crucial case. 

The first $r+1$ ants adopt $\alpha=0$ and make choices at random:
\[
X(k,s)\sim \mbox{Ber}(1/2),k=1,\cdots,N,s=1,\cdots,r+1
\]

We denote the history of the process as $H_t$. 
Here $H_t$ encompasses all choices ${\Delta k^{OUT}(i,s)}$ 
for $s \le t-1$, and ${X(i,s)}$ for $s\le t$. 
Using the information of $H_t$, one can estimate $Z(k,s),k=1,\cdots,N,s\le t$. 
The conditional expected value of $X(i,t+1)$ is:
\[
E[X(k,t+1)|H_t]=f(Z(k,t)).
\]
We also introduce the magnetization $M(i,t)$ as:
\begin{eqnarray}
M(i,t)&\equiv& E[\sigma(i,t+1)|H_t] \nonumber \\
&=&2(f(Z(i,t)))-1 \nonumber \\
&=& 2\alpha(Z(i,t)-\frac{1}{2})\in [-\alpha,\alpha] \nonumber.
\end{eqnarray}
As $X(i,t+1)$ and $X(j,t+1)$ are conditionally independent, the conditional expected value of $E(t+1)$ is estimated as:
\begin{eqnarray}
E[E(t+1)|H_t]&=&-\sum_i h M(i,t) \nonumber \\
&-&\frac{1}{N-1}\sum_{i,j,i\neq j}J M(i,t)M(j,t). \nonumber 
\end{eqnarray}

\section{Dynamics of Pheromone Ratios and Phase Transition}
In this section, we examine the temporal evolution of the system. When ant $t+1$ enters the system, 
$r$ different ants are selected from the preceding $t$ ants, and the out-degrees $\{\Delta k^{OUT}(i,t)\}$ 
are determined. Ant $t+1$ then aggregates the pheromone quantities associated with choice $X(k,s) = x$ and 
makes decisions for $\{X(k,t+1)\}$. Our analysis begins with the time evolution of the network structure.

\subsection{Time Evolution of Network}
The cumulative out-degree $\{k^{OUT}(i,t)\}$ for $t \geq r+1$ is given by:
\[
\sum_{i=1}^{t} k^{OUT}(i,t) = \frac{1}{2}r(r+1) + r(t - (r+1)).
\]
The total popularity $\{l(i,t)\}$ for $t \geq r+2$ is denoted as $D(t)$:
\begin{eqnarray}
D(t) &\equiv& \sum_{i=1}^{t} l(i,t) = rt + \omega \sum_{i=1}^{t} k^{OUT}(i,t) \nonumber \\
&=& rt(1 + \omega) - \omega \frac{1}{2} r(r+1) \nonumber.
\end{eqnarray}
We will hereafter assume $l(i,t) \geq 0$ and omit $\max(l(i,t), 0)$ in the probabilistic rule:
\[
\text{Prob}(\Delta k^{OUT}(i,t)) = \frac{r l(i,t)}{D(t)}.
\]
This assumption is crucial for the computation of the system's time evolution but does not hold when $\omega < 0$. 
Specifically, as $\omega \to -1$, a more meticulous approach is necessary for treating the network's evolution.

\subsection{Mean Field Approximation}
The stochastic process from $t$ to $t+1$ is divided into two stages. The first involves the evolution of 
the network $\{\Delta k^{OUT}(i,t)\}$, and the second pertains to decision-making $\{X(k,t+1)\}$. 
The history of the system, denoted as $H_t$, includes $\{X(i,s)\}, s \leq t$ and $\{\Delta k^{OUT}(i,s)\}, s \leq t-1$. 
Using the information in $H_t$, one can estimate $l(i,s), s \leq t$ and $D(s), s \leq t$. 
The variable $\Delta k^{OUT}(i,t)$ is  a Markov process conditioned on $H_t$. 
We define the conditional expected value of $\Delta k^{OUT}(i,t)$ under $H_t$ as follows:
\begin{eqnarray}
E_{\Delta k}[\Delta k^{OUT}(i,t)|H_t] &=& P(\Delta k^{OUT}(i,t) = 1) \nonumber \\
&=& \frac{r l(i,t)}{D(t)} \nonumber.
\end{eqnarray}
Here, the suffix $\Delta k$ of $E_{\Delta k}$ represents the expected value of 
$\Delta k^{OUT}(t)$. We also express the conditional expected values of $S(t)$ and 
$S_x(k, t)$ as $\tilde{S}(t)$ and $\tilde{S}_x(k, t)$, defined as:
\begin{eqnarray}
\tilde{S}(t) &\equiv& E_{\Delta k}[S(t)|H_t] \nonumber \\
&=& \frac{\sum_{i=1}^{t} r l(i,t) e^{-E(i)}}{D(t)}, \nonumber \\
\tilde{S}_x(k, t) &\equiv& E_{\Delta k}[S_x(k, t)|H_t] \nonumber \\
&=& \frac{\sum_{i=1}^{t} r l(i,t) e^{-E(i)} \delta_{X(k,i),x}}{D(t)} \nonumber.
\end{eqnarray}

$X(k,t+1)$ is modeled as a Markov process conditioned on $H_t$ and $\{\Delta k^{OUT}(i,t)\}$. 
Utilizing this information, one can estimate $S(t)$ and $S_1(k, t)$. We apply the mean field approximation, 
replacing the referencing network $\{\Delta k^{OUT}(i,t)\}$ for ant $t+1$ with the average weighted by 
$rl(i,t)/D(t)$\cite{Hisakado:2023}. Additionally, one can estimate $\tilde{S}(t)$ and $\tilde{S}_1(k, t)$ using 
the information contained in $H_t$. We substitute $Z(k,t) = S_1(k,t)/S(t)$ with 
$\tilde{Z}(k,t) = \tilde{S}_1(k,t)/\tilde{S}(t)$ in eq.(\ref{eq:Px}). 
The pheromone ratio under the mean field approximation is given by:
\[
\tilde{Z}(k, t) = \frac{\tilde{S}_1(k, t)}{\tilde{S}(t)} 
= \frac{\sum_{i=1}^{t} l(i,t) e^{-E(i)} \delta_{X(k,i),1}}{\sum_{i=1}^{t} l(i,t) e^{-E(i)}}.
\]
The probability of making the choice $X(k,t+1)=1$ under $H_t$ in the mean field approximation is expressed as:
\[
P(X(k,t+1) = 1|H_t) = f(\tilde{Z}(k,t)).
\]
Thus, as $\tilde{Z}(k,t)$ can be estimated using the information from $H_t$, 
$X(k,t+1)$ becomes a Markov process conditioned on $H_t$ by the mean field approximation.

\subsection{Recursive Relationship for $\tilde{S}(t)$}
To understand the dynamics of $\{X(k,t)\}$, we estimate the evolution of the pheromone ratios 
$\{\tilde{Z}(k,t)\}=\tilde{S}_1(k,t)/\tilde{S}(t)$\cite{Mori:2024}. 
We begin by deriving the recursive relationship for $\tilde{S}(t)$:
\[
\tilde{S}(t+1) = \frac{r}{D(t+1)}\sum^{t+1}_{i=1}l(i,t+1)e^{-E(i)}
\]
The recursive relation for popularity is given by:
\[
l(i,t+1)=l(i,t) + \omega \Delta k^{OUT}(i,t).
\]
Applying the mean field approximation, we replace $\Delta k^{OUT}(i,t)$ 
with $\frac{rl(i,t)}{D(t)}$, thus:
\[
l(i,t+1)=(1+\frac{\omega r}{D(t)})l(i,t).
\]
The evolution of $\tilde{S}(t)$ is then:
\begin{eqnarray}
\tilde{S}(t+1) &=& \frac{r}{D(t+1)} \left\{ \sum_{i=1}^{t}\left(1+\frac{\omega r}{D(t)}\right)l(i,t)e^{-E(i)} \right. \nonumber \\
&+& \left. l(t+1,t+1)e^{-E(t+1)} \right\} \nonumber
\end{eqnarray}
Since $D(t+1)=D(t)+r(1+\omega)$ and $l(t+1,t+1)=r$, we find:
\begin{eqnarray}
\tilde{S}(t+1) &=& \frac{1}{D(t+1)}\left\{(D(t)+\omega r)\tilde{S}(t) \right. \nonumber \\
&+& \left. r^2 e^{-E(t+1)}\right\} \nonumber \\
&=& \tilde{S}(t)+\frac{r}{D(t+1)}(re^{-E(t+1)}-\tilde{S}(t)) \nonumber
\end{eqnarray}
We thus obtain the recursive relation for $\tilde{S}(t)$:
\[
\Delta \tilde{S}(t)=\frac{r}{D(t+1)}(re^{-E(t+1)}-\tilde{S}(t)).
\]
$\tilde{S}(t)$ is a stochastic process, and its fluctuation originates from the conditional 
variance of $e^{-E(t+1)}$ under $H_t$. Fluctuations are neglected and $e^{-E(t+1)}$ is replaced by 
its conditional expected value. In the continuous time limit, the differential equation for $\tilde{S}(t)$ is derived as:
\[
\frac{d}{dt}\tilde{S}(t)=\frac{r}{D(t+1)}(E[re^{-E(t+1)}|H_t]-\tilde{S}(t)).
\]
In the stationary state, $\tilde{S}(t)$ converges to:
\[
\tilde{S}(t) \to E_{st}[re^{-E(t)}],
\]
where $E_{st}[*]$ indicates the expected value in the stationary state.

\subsection{Recursive Relationship for $\tilde{S}_1(k,t)$}
Next, we develop the recursive relation for $\tilde{S}_1(k,t)$. 
We begin with the expression for $\tilde{S}_1(k,t+1)$:
\begin{eqnarray}
&&\tilde{S}_1(k, t+1) \nonumber  \\
&=& \frac{r}{D(t+1)}\sum^{t+1}_{i=1}l(i, t+1)e^{-E(i)}\delta_{X(k, i),1} \nonumber
\end{eqnarray}
Following the similar procedure used for $\tilde{S}(t+1)$, we find:
\begin{eqnarray}
&&\Delta \tilde{S}_1(k,t) \nonumber \\
&=&\frac{r}{D(t+1)}(re^{-E(t+1)}\delta_{X(k,t+1),1}-\tilde{S}_1(k,t)) \nonumber .
\end{eqnarray}
As in our previous work, we expand $E(t+1)\delta_{X(k,t+1),x}$:
\begin{eqnarray}
&&E(t+1)\delta_{X(k,t+1),x}\nonumber \\
&=&E(t+1)+\frac{\partial E(t+1)}{\partial X(k,t+1)}(x-X(k,t+1))\nonumber .
\end{eqnarray}
We then introduce the "effective field" $\tilde{h}(k, t)$, defined by:
\begin{eqnarray}
\tilde{h}(k,t) &=& h+\frac{1}{N-1}\sum_{l \neq k}2J(2X(l,t)-1) \nonumber \\
&=&-\frac{1}{2} \frac{\partial E(t)}{\partial X(k,t)}. \nonumber
\end{eqnarray}
Assuming $\tilde{h}(k, t+1)$ is small, which holds true for $h \ll 1$ and $J \ll 1$, the expansion becomes:
\begin{eqnarray}
&&E(t+1)\delta_{X(k,t+1),x}\nonumber \\
&=&E(t+1)-2\tilde{h}(k,t+1)(x-X(k,t+1)) \nonumber.
\end{eqnarray}
Applying this expansion to $e^{-E(t+1)\delta_{X(k,t+1),1}}$, we approximate:
\begin{eqnarray}
&&e^{-E(t+1)\delta_{X(k,t+1),1}}\nonumber \\
&\simeq& e^{-E(t+1)+2\tilde{h}(k,t+1)(1-X(k,t+1))}\nonumber \\
&\simeq& e^{-E(t+1)}(1+2\tilde{h}(k,t+1)(1-X(k,t+1)))\nonumber.
\end{eqnarray}
Consequently, we derive:
\begin{eqnarray}
&&\Delta \tilde{S_1}(k,t)=\frac{r}{D(t+1)}(re^{-E(t+1)}
\nonumber \\
&&\{1+2\tilde{h}(k,t+1)(1-X(k,t+1))\}\delta_{X(k,t+1),1}
\nonumber \\
&&-\tilde{S}_1(k,t)) \nonumber .
\end{eqnarray}

\subsection{Stochastic differential equation for $\tilde{Z}(k, t)$}
Finally, we derive the recursive relation for $\tilde{Z}(k, t)$. 
The relationship is given by:
\[
\tilde{Z}(k,t+1)=\frac{\tilde{S}_1(k,t+1)}{\tilde{S}(t+1)}=
\frac{\tilde{S}(t)}{\tilde{S}(t+1)}\cdot \frac{\tilde{S}_1(k,t+1)}{\tilde{S}(t)}
\]
Considering $\tilde{S}_1(k,t+1) = \tilde{S}_1(k,t) + \Delta \tilde{S}_1(k,t)$, we deduce:
\begin{eqnarray}
&&\tilde{Z}(k,t+1)=\frac{\tilde{S}(t)}{\tilde{S}(t+1)} \nonumber \\
&&(\tilde{Z}(k,t)+\frac{r}{D(t+1)}[\frac{re^{-E(t+1)}}{\tilde{S}(t)}\{1-2\tilde{h}(k,t+1)
\nonumber \\
&&(x-X(k,t+1))\}\delta_{X(k,t+1),1}-\tilde{Z}(k,t)]
)
\end{eqnarray}
In the stationary state where $\tilde{S}(t) = \tilde{S}(t+1) \approx E_{st}[re^{-E(t)}]$, the change in $\tilde{Z}(k,t)$ is:
\begin{eqnarray}
&&\Delta \tilde{Z}(k,t+1)=\tilde{Z}(k,t+1)-\tilde{Z}(k,t)\nonumber \\ 
&=&\frac{r}{D(t+1)}(\{1-2\tilde{h}(k,t+1)(1-X(k,t+1))\}\nonumber \\
&&\delta_{X(k,t+1),1}-\tilde{Z}(k,t))
\end{eqnarray}

We derive stochastic differential equations (SDE). 
The expected value and variance of $\Delta \tilde{Z}(k, t)$, conditioned on the history $H_t$, are estimated as follows:
\begin{eqnarray}
&&E[\Delta\tilde{Z}(k,t)|H_t]=\frac{r}{D(t+1)}\{E[(1+ 
\nonumber \\
&&2\tilde{h}(k,t+1)(1-X(k,t+1)))\delta_{X(k, t+1),1}|H_t] 
\nonumber \\
&&-\tilde{Z}(k, t)\} \nonumber
\end{eqnarray}
We approximate the expected value of the product of random variables by the product of their expected values:
\begin{eqnarray}
&&E[\Delta\tilde{Z}(k,t)|H_t]\simeq \frac{r}{D(t+1)}\{f(\tilde{Z}(k,t))\nonumber \\
&&[1+2E[\tilde{h}(k,t+1)|H_t](1-f(\tilde{Z}(k, t)))] \nonumber \\
&&-\tilde{Z}(k, t)\}, \nonumber \\
&&E[\tilde{h}(k,t+1)|H_t]\nonumber \\
&& =h+\frac{2J}{N-1}\sum_{l\neq k}(2f(\tilde{Z}(l,t))-1) \nonumber .
\end{eqnarray}
The conditional variance is approximated by:
\begin{eqnarray}
&&V[\Delta \tilde{Z}(k, t)|H_t] \nonumber \\
&&\simeq \frac{r^2}{D(t+1)^2}f(\tilde{Z}(k, t))(1-f(\tilde{Z}(k, t))) \nonumber .
\end{eqnarray}
Finally, the SDEs for $\{Z(k,t)\}$ are expressed as:
\begin{eqnarray}
&&d\tilde{Z}(k, t)= E[\Delta \tilde{Z}(k, t)|H_t]dt  \nonumber \\
&&+\sqrt{V[\Delta \tilde{Z}(k, t)|H_t]}dW(k, t) 
\label{eq:SDE}.
\end{eqnarray}
Here, $\vec{W}(t) = \{W(k,t)\}$, $k=1, \ldots, N$, represents an array of independent and identically distributed 
Wiener processes, and $d\vec{W}(t)$ follows a $N_{N}(0, I dt)$ distribution, denoting the $d$-dimensional normal 
distribution with mean $\vec{\mu}$ and variance $\Sigma$ as $N_{d}(\vec{\mu}, \Sigma)$.

\subsection{$t\to\infty$ limit of the Pheromone ratios}
To clarify the structure of the SDEs, we approximate equation (\ref{eq:SDE}). 
Assuming $f(z) = \frac{1}{2} + \alpha(z - \frac{1}{2})$ and 
$1-f(z) = \frac{1}{2} - \alpha(z - \frac{1}{2})$ for small $\alpha$, we derive:
\[
f(z)(1-f(z))=\frac{1}{4}-\alpha^2\left(z-\frac{1}{2}
\right)^2\simeq \frac{1}{4}.
\]
Additionally, with $f(z) - z = (1 - \alpha)(\frac{1}{2} - z)$, the SDEs in eq.(\ref{eq:SDE}) simplify to:
\begin{eqnarray}
&&d\tilde{Z}(k,t)=\frac{r}{D(t)}\{-(1-\alpha)(\tilde{Z}(k,t)-1/2) \nonumber \\
&+&\frac{1}{2}(h+\frac{2J}{N-1}\sum_{l\neq k}2\alpha(\tilde{Z}(l,t)-1/2))\}dt \nonumber \\
&+&\frac{r}{2D(t)}dW(k,t) \label{eq:SDE2}.
\end{eqnarray}

\begin{figure}[htbp]
\begin{center}
\includegraphics[width=5cm]{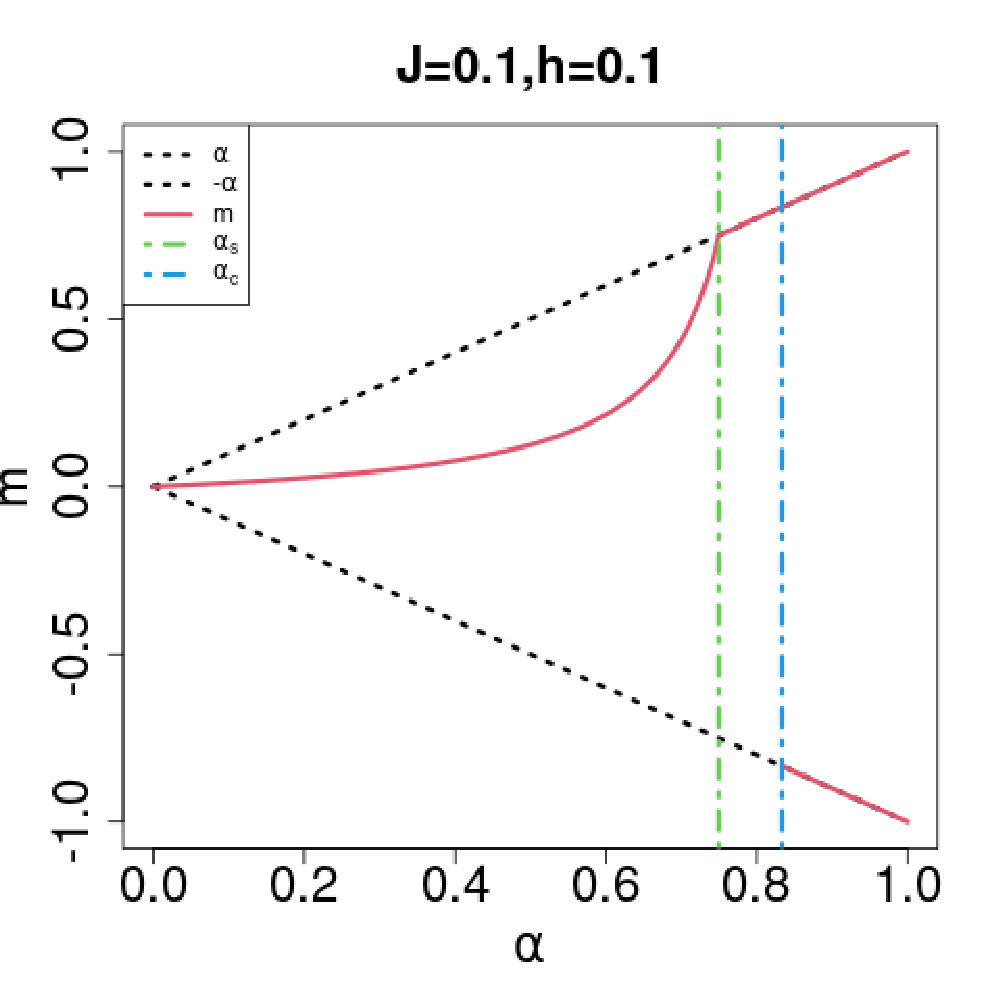}
\caption{\label{fig3}Plot of $m_*$ vs. $\alpha$ for $J=h=0.1$. 
The vertical lines show the position of $\alpha_s$ and $\alpha_c$.}
\end{center}
\end{figure}

We express the SDEs in terms of the magnetization $M(k,t) = 2\alpha(\tilde{Z}(k,t)-\frac{1}{2}) 
\in [-\alpha, \alpha]$. Multiplying both sides of eq.(\ref{eq:SDE2}) by $2\alpha$, we obtain:
\begin{eqnarray}
&&dM(k,t)=\frac{r}{D(t)}\cdot \nonumber \\
&&\{-(1-\alpha)M(k,t)+\alpha(h+\frac{2J}{N-1}\sum_{l\neq k}M(l,t))\}dt \nonumber \\
&&+\frac{\alpha r}{D(t)}dW(k,t) \label{eq:SDE3}.
\end{eqnarray}
We introduce a potential $U(\{M(k)\})$ defined as:
\begin{eqnarray}
&&U(\{M(k)\})\equiv \frac{1}{2}(1-\alpha)\sum_{k}M(k)^2 \nonumber \\
&-&\alpha \{h\sum_{k}M(k)+\frac{J}{N-1}\sum_{k\neq l}M(k)M(l)\}.\nonumber 
\end{eqnarray}
The SDEs in eq.(\ref{eq:SDE3}) are then represented as:
\begin{eqnarray}
dM(k,t)&=&-\frac{r}{D(t)}\nabla_{M(k,t)}U(\{M(k,t)\}) \nonumber \\ 
&+&\frac{\alpha r}{D(t)}dW(k,t)\label{eq:SDE4}.
\end{eqnarray}
As the potential $U(\{M(k,t)\})$ consists of linear and quadratic terms in $\{M(k,t)\}$, 
the SDEs describe a multi-variate Ornstein-Uhlenbeck process \cite{Gardiner:2009,Mori:2024}. 
Generally, $M(k,t)$ is driven towards the minimum of the potential. Given uniform interactions $J$ and 
external field $h$, the minimum of the potential does not depend on $k$. The potential can then be expressed as:
\begin{eqnarray}
&&u(m)\equiv U(\{M(k)\}=m)/N \nonumber \\
&=&\frac{1}{2}(1-\alpha(1+2J))m^2-\alpha h m \nonumber .
\end{eqnarray}
The minimum $m_{*}$ of $u(m)$ is derived as,
\begin{equation}
m_{*}=
\left\{
\begin{array}{cc}
\frac{\alpha h}{1-\alpha(1+2J)} & \alpha<\alpha_s=\frac{1-h/2}{1+2J} \\
\alpha & \alpha\ge \alpha_s
\end{array}
\right. \label{eq:m_st}
\end{equation}
However, $m_*$ becomes unstable when $\alpha > \alpha_c = \frac{1}{1+2J}$, as the sign of 
the quadratic term in $u(m)$ turns negative. In such cases, $m = \pm 1$ become stable. 
Figure \ref{fig3} summarizes the results for $m_{*}$ for the case $J = 0.1$ and $h = 0.1$.

The common factor $D(t)$ in the denominators of the drift and diffusion terms in eq.(\ref{eq:SDE}) 
plays a crucial role in the dynamics of $M(k,t)$. In the $\omega \to -1$ limit, the network becomes 
an extended lattice and $D(t) = \frac{r(r+1)}{2}$. When pheromones evaporate with a time scale $\tau$, 
similar SDEs are obtained by replacing $D(t)/r$ with $\tau$ \cite{Mori:2024}. The extended lattice case 
corresponds to the typical ACO system. The stationary distribution of $\{M(k,t)\}$ is given by:
\begin{equation}
p_{st}(\{m(k)\})\propto e^{-U(\{m(k)\})/(2\alpha^2/(r+1)^2)}.
\end{equation}
$\{M(k,t)\}$ fluctuate around the minimum of the potential $U$, if the minimum exists within the range $[-\alpha, \alpha]$. 
The parameter $r$ determines the variance of the distribution. When $\omega > -1$, $D(t) \propto (1+\omega)rt$ and 
diverges as $t \to \infty$, reducing the influence of the drift and diffusion terms. This situation is akin to the 
case where $\tau \to \infty$, and pheromones do not evaporate. Here, $M(k,t)$ generally converges to the minimum 
$m_{*}$ of $u(m)$. However, when $\omega$ is large, the drift terms diminish quickly, and $M(k,t)$ may not have 
sufficient time to approach $m_{*}$. When $\omega \sim -1$, $D(t)$ remains small for an extended period, 
allowing $M(k,t)$ ample time to approach $m_{*}$.

For $\alpha < \alpha_c$, for large $T$, $M(k,T)$ is a monotonically decreasing function of $\omega$. 
When $\omega \simeq -1$, $M(k,T)$ is maximal. Conversely, when $\omega \gg 1$, $M(k,T)$ remains near zero. 
In the limit as $\omega \to -1$ (the extended lattice case), $M(k,T)$ is distributed around $m_{*}$. 
For $\alpha > \alpha_c$, two stable states $\pm \alpha$ exist, and it becomes random as to which state $M(k,t)$ converges. 
The average value of $M(k,T)$ over many sample paths is small due to the discontinuous bifurcation transition. 

\section{Numerical Studies}
We evaluated the theoretical results from the previous section using Monte Carlo simulations. 
Magnetization $\{M(k,t)\}$ was sampled for each combination of $\omega$ and $\alpha$, 
where $\omega \in \{1.0, 0.0, -0.9999, -1.0\}$ and $\alpha \in [0.5, 0.99]$. 
We utilized $N=100$ spins and performed $S=100$ trials, each consisting of $T=10^5$ ants. 
Each value of $\omega$ was tested with one network sample. The parameters of the Ising model were 
set to $h = 0.01$ and $J = 0.1$, and the number of in-degrees, $r$, was set to $r = 10^2$. 
The magnetization $M(k, t, s)$ represents the magnetization of ant $t$ for spin $k$ 
during trial $s$.

\begin{figure}[h]
    \centering
    \includegraphics[width=6cm]{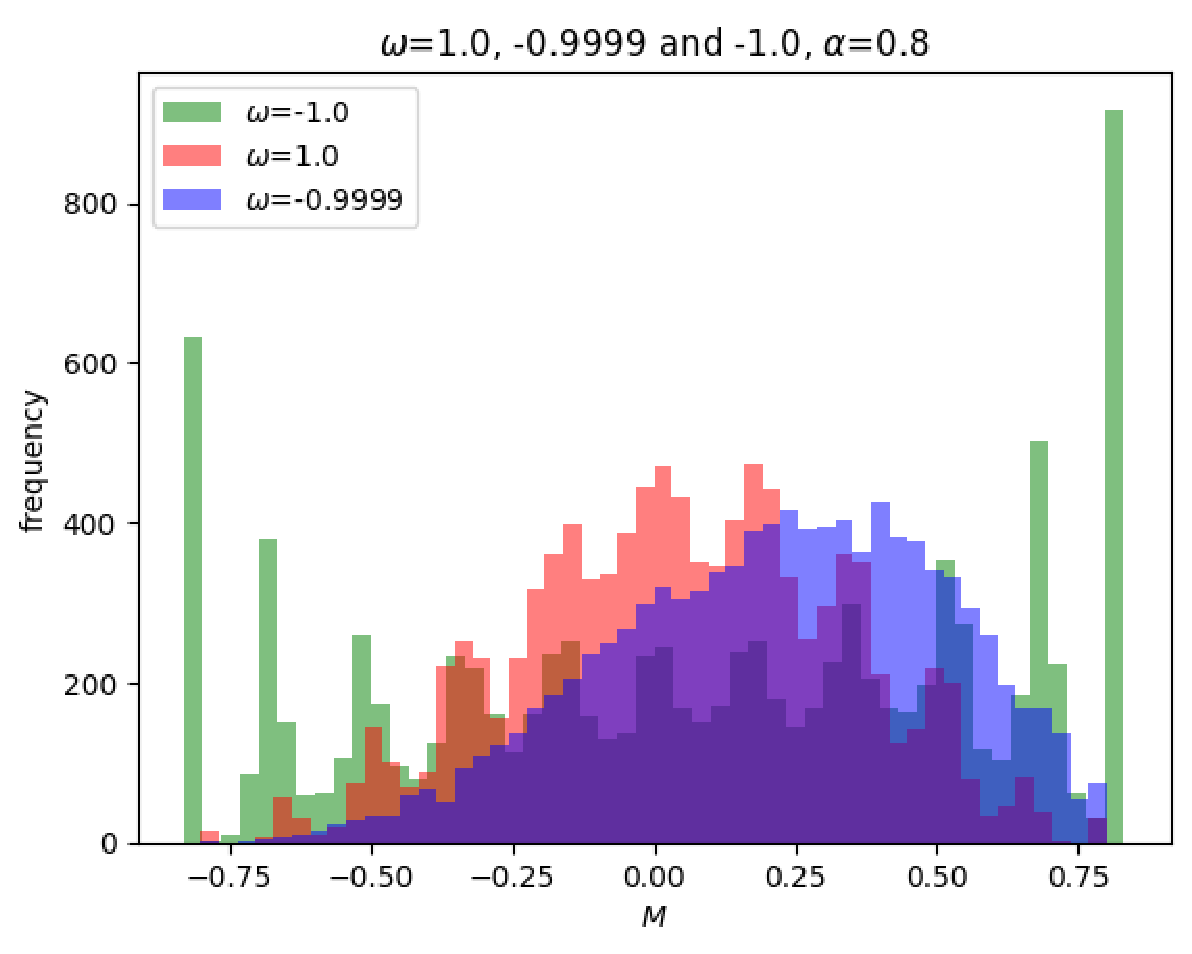}
    \caption{Histograms of $M(k,T,s)$ for $\alpha=0.8$ and $\omega \in \{-1.0, -0.9999, 1.0\}$.}
    \label{fig:fig4}
\end{figure}
Figure \ref{fig:fig4} displays the histograms of $M(k,T,s)$ for $\alpha=0.8$ and $\omega \in \{-1.0, -0.9999, 1.0\}$. 
A comparison between the histograms for $\omega=-0.9999$ (blue) and $\omega=1.0$ (red) shows that $M(k,t,s)$ is effectively 
influenced by $\tilde{h}(k,t)$ at $\omega=-0.9999$. Conversely, when $\omega=1.0$, the drift term does not effectively 
influence $M(k,t,s)$, which remains around $0.5$. The histogram for $\omega=-1.0$ differs markedly from the other two, 
suggesting that mean field treatments are invalid at $\omega=-1.0$. Alternative methods should be considered for 
the extended lattice case.

\begin{figure}[h]
    \centering
    \includegraphics[width=6cm]{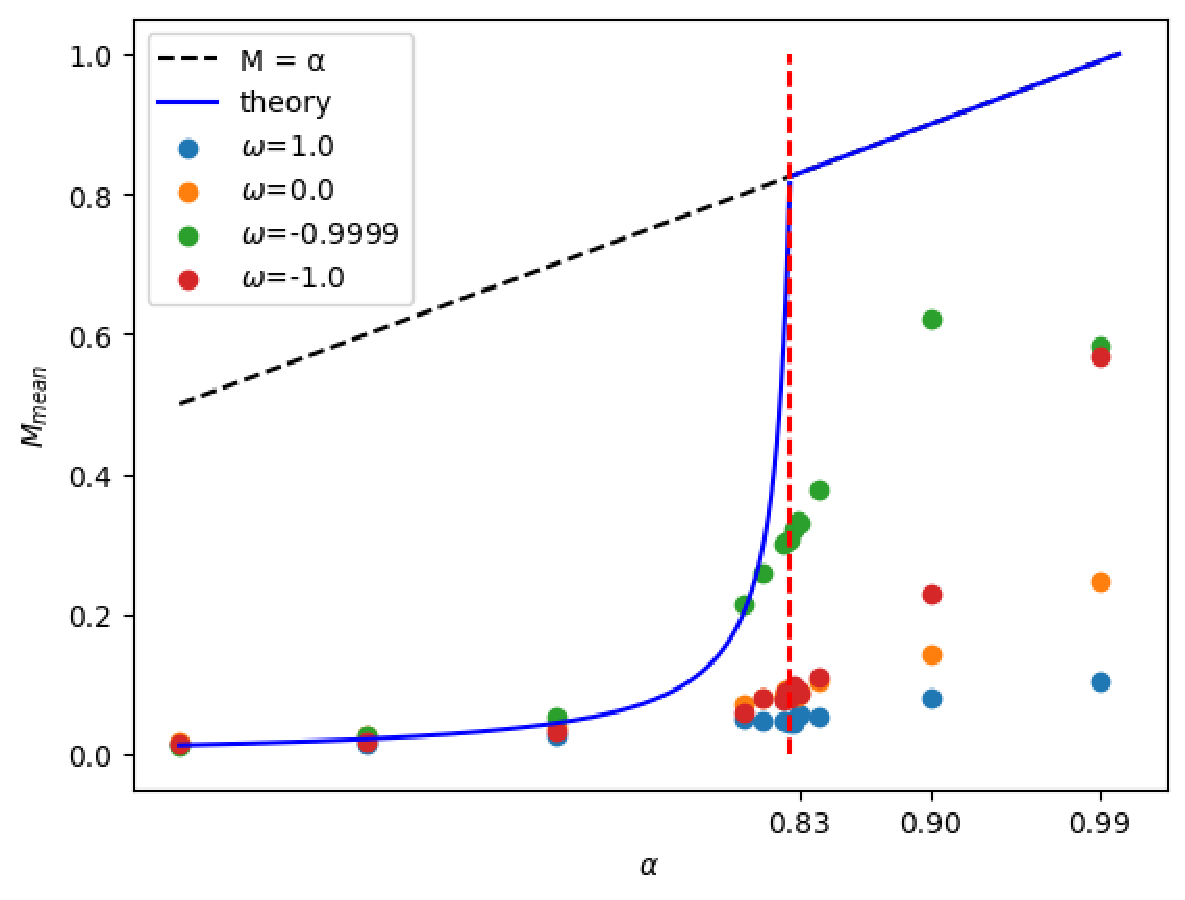}
    \caption{Plot of $M_{mean}$ versus $\alpha$ for $\omega \in \{-1.0, -0.9999, 0.0, 1.0\}$. 
    The blue solid line represents $m_{*}$ from equation (\ref{eq:m_st}), and the red dashed line 
    indicates $\alpha_c=1/(1+2J)$.}
    \label{fig:fig5}
\end{figure}

Figure \ref{fig:fig5} presents the mean magnetization $M_{mean}$, calculated as:
\[
M_{mean}=\frac{1}{NS}\sum_{k,s}M(k,T,s).
\]
For each $\omega$, $M_{mean}$ versus $\alpha$ is plotted. At $\omega=-0.9999$, $M_{mean}$ closely aligns with $m_{*}$ (blue line). 
However, for $\omega=0.0$ and $\omega=1.0$, $M_{mean}$ approximates $0.5$, suggesting the absence of significant effective field influence, 
as theoretically anticipated from the behavior of $D(t)$.

To assess the ability to find the ground state, we analyze the data $\{M(k,T,s)\}$ to determine $\sigma(k,s)$ as follows:
\[
\sigma(k,s)=\sgn(M(k,T,s)),
\]
where $\sgn(x)=\pm 1$ for $x \geq 0$ and $x < 0$, respectively. The system is considered to have found the ground state 
when all $N$ $\sigma(k,s)$ values are 1. The success probability is estimated by:
\begin{equation}
\mbox{Success Probability} = \frac{1}{S}\sum_{s=1}^{S} \prod_{k=1}^{N} \delta_{\sigma(k,s),1}
\end{equation}

\begin{figure}[h]
    \centering
    \includegraphics[width=6cm]{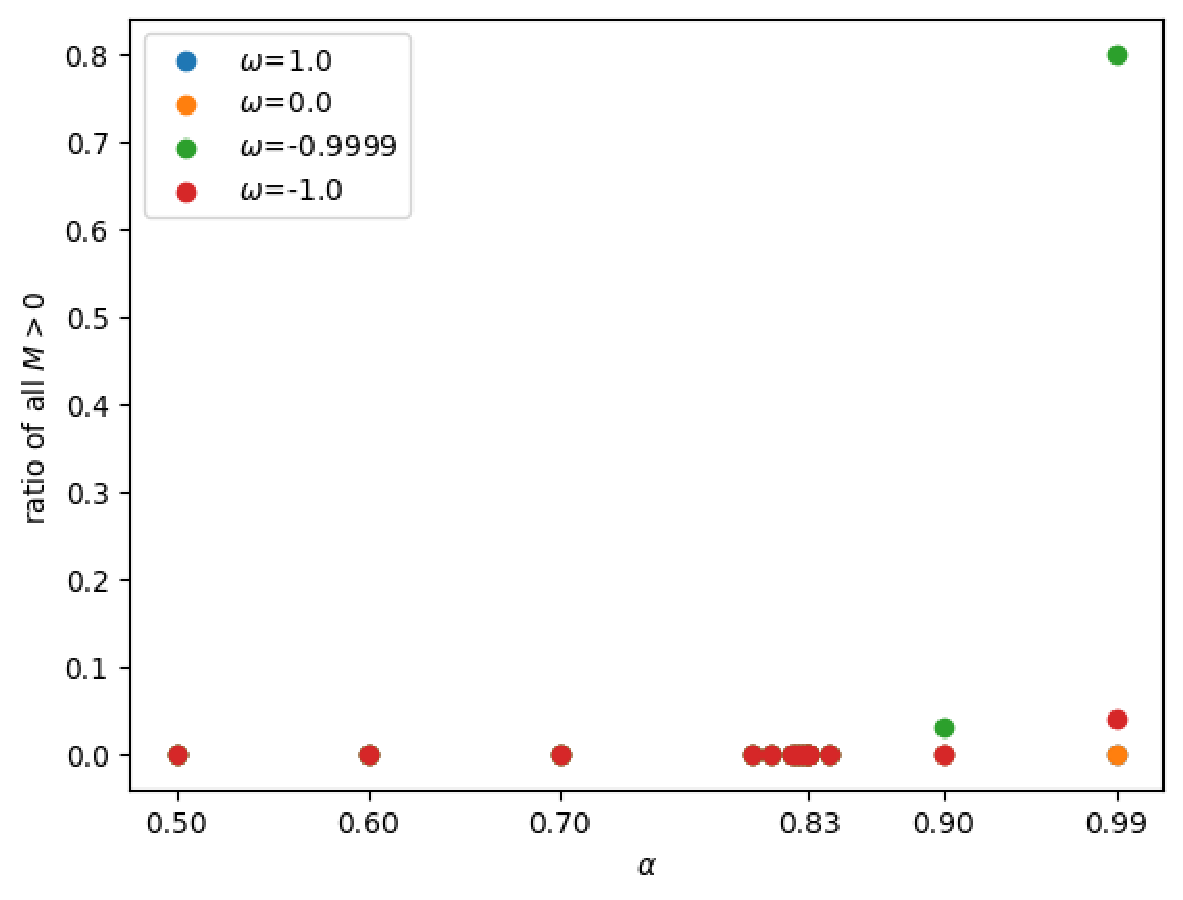}
    \caption{Success probability versus $\alpha$, testing $\omega \in \{-1.0, -0.9999, 0.0, 1.0\}$ and $N=100$.}
    \label{fig:fig6}
\end{figure}

Figure \ref{fig:fig6} illustrates the success probability as a function of $\alpha$. 
At $\omega=-0.9999$, the success probability increases sharply from nearly 0 at $\alpha=0.83$ to about 0.8 at $\alpha=0.99$. 
To align all spins uniformly, $\alpha \approx 1$ is necessary. In contrast, for $\omega=1.0, 0.0, -1.0$, the success probability is 
almost zero, indicating that the system fails to find the ground state.

\section{Conclusion}
In this paper, we have studied the impact of network structure on the performance of Ant Colony Optimization (ACO). 
In traditional ACO models, pheromones evaporate, and ants refer information from recent predecessors in chronological order. 
Typically, ACO employs an extended lattice, characterized by a large average node-to-node distance. We introduce a network 
model for ACO where ants select $r$ different ants based on their popularity, controlled by the parameter $\omega$: at $\omega=-1$, 
it resembles an extended lattice; at $\omega=0$, a random graph; and at $\omega=1$, a scale-free network with prominent hubs.

Ants in our model search for the ground state of the infinite-range Ising model, with pheromone amounts determined by the Boltzmann weight 
of the solution. The decision function of ants is linear, with their response to pheromones described by the parameter $\alpha$. 
We analyzed the dynamics of the pheromone ratio using stochastic differential equations and derived the system's phase diagram, 
which exhibits a discontinuous bifurcation transition when $\alpha$ exceeds a critical value $\alpha_c$. 
These theoretical results were validated through Monte Carlo simulations.

We report three new findings:
1. The discontinuous bifurcation transition triggered by changes in $\alpha$ for $\omega > -1$ extends our previous findings, 
where no evaporation of pheromones occurred\cite{Mori:2024}. This suggests that network structure significantly restricts the ants’ 
pheromone reference network, allowing older pheromone information to persist longer. Consequently, for large $\alpha$, the system may exhibit two stable states.
2. The pheromone ratios converge rapidly to the stable state as $\omega \to -1$, facilitated by the common factor $D(t)$ in the drift and diffusion term denominators.
3. The probability of finding the ground state is maximal when $\omega \to -1$. To align all variables $\{M(k,t)\}$, it is necessary to use $\alpha \to 1$, 
which exceeds $\alpha_c$. It remains random to which stable state $m = \pm 1$ the system will converge, yet the probability remains at its peak.

In previous work, we demonstrated that an $\alpha$-annealing process, which gradually increases $\alpha$ to maintain the system in an optimal state, 
is effective for searching the ground state of the mean-field Ising model\cite{Mori:2024-2}. From this viewpoint, the discontinuous bifurcation transition 
should be avoided. We believe the evolution of the pheromone reference network should consider the aging effect, whereby the popularity of ants 
decays over time. Determining the optimal network structure for this scenario remains an open question for future research.

\section*{Acknowledgements}
The authors would like to thank Mr. Shogo Nakamura for his valuable discussions and for providing access to the 
Monte-Carlo simulation program. This work was supported by the Japan Society for the Promotion of Science (JSPS) KAKENHI, 
Grant Number 22K03445.


\end{document}